\documentclass[10pt,twocolumn,letterpaper]{article}
\newcommand{\ARXIV}[2]{#2} %
\usepackage{templates/EV/cvpr}              %

\definecolor{softblue}{rgb}{0.12,0.49,0.85}
\usepackage[pagebackref,breaklinks,colorlinks,citecolor=softblue,hyperfootnotes=false]{hyperref}

\def\paperID{12} %
\def\confName{EarthVision}
\def\confYear{2026}

\PassOptionsToPackage{hyphens}{url}

\usepackage{graphicx}
\usepackage{amsmath}
\usepackage[table]{xcolor}
\usepackage{amssymb}
\usepackage{times}
\usepackage{mathtools}
\usepackage{amssymb}
\usepackage{booktabs}
\usepackage{multirow,multicol}
\usepackage{tabularx}
\usepackage{xcolor}
\usepackage{balance}
\usepackage{placeins}
\usepackage{subcaption}
\usepackage{enumitem}
\usepackage{tikz}
\usepackage{pgfplots}
\usepackage{styles}
\usepackage{makecell}
\usepackage{microtype}
\usepackage{paralist}
\usepackage{tcolorbox}
\usepackage{siunitx}
\newsavebox{\tempbox}

\usetikzlibrary{shapes, shapes.geometric, shadows, arrows.meta, calc, decorations.pathreplacing, fadings, positioning}
\usepackage[hang,flushmargin]{footmisc}
\usepackage[accsupp]{axessibility}  %

\renewcommand{\thefootnote}{\fnsymbol{footnote}}

\begin{document}

\title{\TITLE}

\author{
Guillaume Astruc\textsuperscript{1,3,4,}\footnotemark\;,\quad\;
Eduard Trulls\textsuperscript{2}\quad\; 
Jan Hosang\textsuperscript{2}\quad\; 
Loic Landrieu\textsuperscript{1,4}\quad\; 
Paul-Edouard Sarlin\textsuperscript{2}
\and
{
\textsuperscript{1} LASTIG, Univ Gustave Eiffel, IGN, ENSG, France \quad\;
\textsuperscript{2} Google, Switzerland
\quad\;
\textsuperscript{3} CNES, France
}
\and{
\textsuperscript{4} LIGM, CNRS, Univ Gustave Eiffel, ENPC, Institut Polytechnique de Paris, Marne-la-Vallée, France
}
}

\maketitle
\begin{abstract}
The growing availability of co-located geospatial data spanning aerial imagery, street-level views, elevation models, text, and geographic coordinates offers a unique opportunity for multimodal representation learning.
We introduce \ours, a massively multimodal contrastive framework to jointly align five complementary geospatial modalities in a single unified embedding space.
Unlike prior approaches that fuse modalities or rely on a central pivot representation, our method performs all-to-all contrastive alignment, enabling seamless comparison, retrieval, and reasoning across arbitrary combinations of modalities.
We further propose a scaled latitude-longitude encoder that improves spatial representation by capturing multi-scale geographic structure.
Extensive experiments across downstream geospatial tasks demonstrate that \ours\ consistently outperforms single-modality contrastive models and coordinate-only baselines, highlighting the benefits of holistic multimodal geospatial alignment.
A reference implementation is available at
\href{https://gastruc.github.io/unigeoclip}{\nolinkurl{gastruc.github.io/unigeoclip}}.

\end{abstract}
\setlength{\parskip}{-0.13em}

\footnotetext[1]{Work done during an internship at Google.}
\renewcommand{\thefootnote}{\arabic{footnote}}
\section{Introduction}
Expressive and robust geospatial embeddings that capture both fine-grained semantic content and large-scale spatial structure are critical for automating downstream geospatial tasks such as urban land-use classification~\cite{lacoste2023geo}, monitoring land cover~\cite{hu2023mdas}, and large-scale socio-economic inference~\cite{sun2024community}.
Existing work largely falls into three successful paradigms.
\emph{Embedding fields} map geographic coordinates to latent vectors to enable localized interpolation~\cite{brown2025alphaearth, lindenberger2025scaling}.
\emph{Multimodal vision models} fuse multiple sensor observations into a single representation~\cite{astruc2024omnisat, astruc2024anysat, tseng2025galileo}.
Finally, \emph{contrastive approaches} align heterogeneous geospatial data sources in a shared embedding space, most notably GeoCLIP \cite{vivanco2023geoclip} and SatCLIP~\cite{klemmer2025satclip}.
Despite their success, these paradigms exhibit limitations for general-purpose geospatial reasoning.
Embedding fields provide static ``snapshots'' of a region and struggle to model dynamic.
Multimodal fusion models collapse all available modalities into a single representation, preventing cross-modal comparison or retrieval.
Existing contrastive approaches typically align geographic coordinates with a single Earth observation modality, most often top-down imagery, and largely ignore textual information despite its central role in modern vision-language models.

\begin{figure}
    \centering
    \resizebox{1\columnwidth}{!}{%
        \input{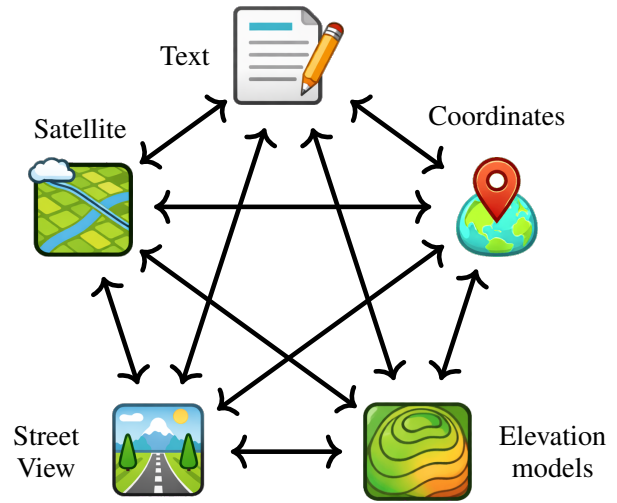}%
    }
\caption{{\bf Unified contrastive learning of geospatial data.}
We jointly train encoders for five modalities (text, aerial imagery, street-level imagery, elevation, and geographic coordinates), which simultaneously are contrasted across all modality pairs.
This yields a single unified embedding space that represents heterogeneous geospatial information.}
\label{fig:teaser}
\end{figure}

\begin{figure*}
    \centering
    \begin{tabular}{c@{\;}c@{\;}c@{\;}c}
\begin{subfigure}{0.25\linewidth}
    \includegraphics[width=\linewidth,height=\linewidth]{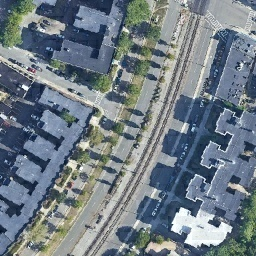}
    \caption{\sat\ Aerial Image.}
\end{subfigure}
&
\begin{subfigure}{0.25\linewidth}
    \includegraphics[width=\linewidth,height=\linewidth]{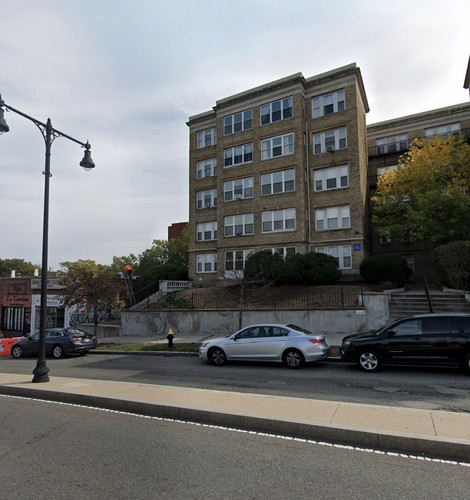}
    \caption{\street\ Street View.%
    }
\end{subfigure}
&
\begin{subfigure}{0.25\linewidth}
    \includegraphics[width=\linewidth,height=\linewidth]{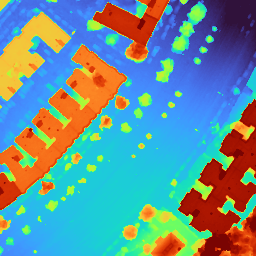}
    \caption{\dsm\ Surface Model.}
\end{subfigure}
&
\begin{subfigure}{0.25\linewidth}
    \includegraphics[width=\linewidth,height=\linewidth,keepaspectratio=true,trim={2mm 2mm 2mm 2mm},clip]{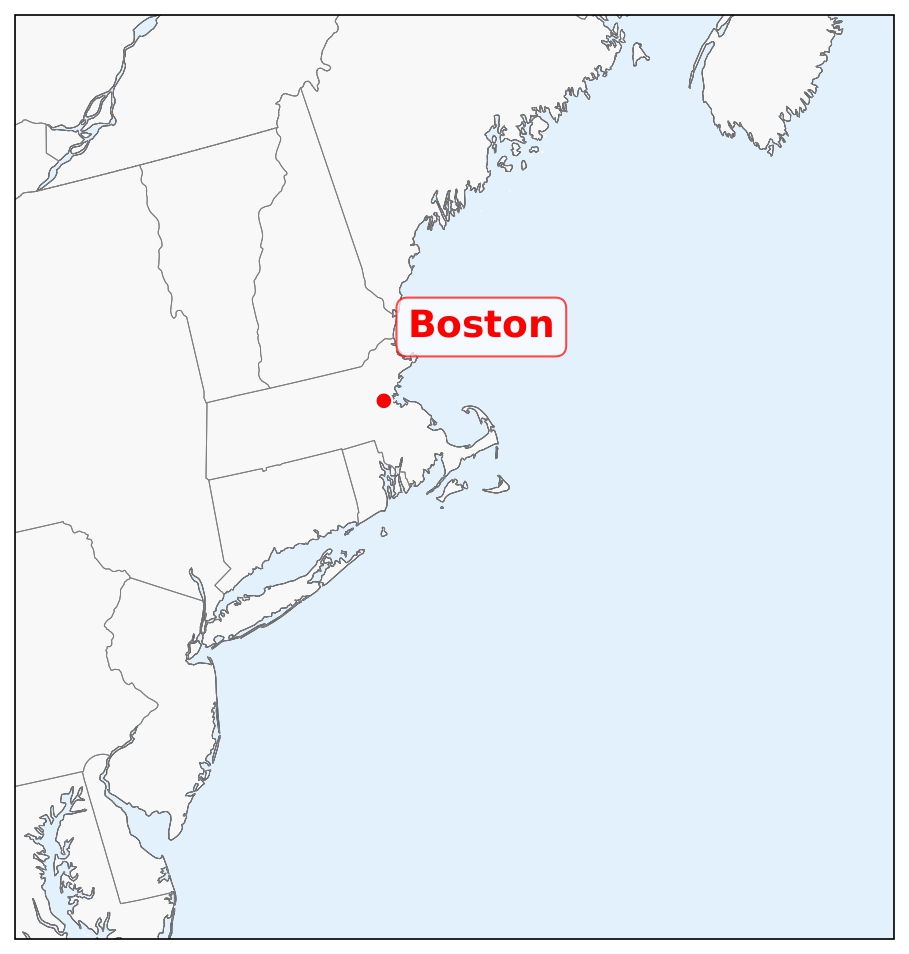}
    \caption{\gps\ Location.}
\end{subfigure}
\\
\multicolumn{4}{c}{
    \begin{subfigure}{1\linewidth}
    \begin{tcolorbox}[
        colback=gray!5!white,
        colframe=gray!50!black,
        boxrule=0.5pt,
        arc=3pt, %
        left=2pt, right=2pt, top=2pt, bottom=2pt %
    ]
        \footnotesize\textit{Situated in Boston's lively Brighton neighborhood, this area is a convenient urban base close to Boston College and with accessible parks. Everyday conveniences include multiple grocery options, including Whole Foods Market and Star Market, as well as diverse restaurants ranging from Spanish tapas at Barcelona Wine Bar to Korean BBQ at Naksan. For nightlife, there's the welcoming atmosphere at Harry's Bar \& Grill and The Publick House, and late-night dining at Barcelona Wine Bar.}
    \end{tcolorbox}%
    \vspace{-0.5em}
    \caption{\textual\ Text Description.}
    \end{subfigure}
}%
\end{tabular}

    \caption{{\bf Sample from our multimodal geospatial dataset.}
Each location is represented through five complementary modalities: aerial imagery, street-level imagery, a Digital Surface Model (DSM), geographic coordinates, and an automatically generated text description. \protect\footnotemark
All modalities are spatially co-registered and jointly contrasted during training.}
    \label{fig:dataset}
\end{figure*}
In this paper, we propose a \emph{contrastive and massively multimodal} framework that jointly aligns five complementary geospatial modalities, enabling seamless translation, comparison, and retrieval across modalities.
Specifically, we contrast high-resolution aerial imagery, geometry-dense Digital Surface Models (DSMs), street-level imagery, rich text descriptions, and raw geographic coordinates, embedded using a novel spatial encoder.
Unlike prior multimodal contrastive frameworks in vision, such as UniBind \cite{lyu2024unibind} or ImageBind  \cite{girdhar2023imagebind}, which rely on a central pivot modality (typically images), our approach adopts a fully holistic formulation: all modalities are directly contrasted with one another.
This all-to-all alignment strategy yields a unified embedding space that supports robust reasoning under arbitrary availability of modalities. Observing that embeddings derived from raw positional encodings are often limited in expressiveness~\cite{russwurm2023geographic} and can become a bottleneck when contrasted with richer modalities, we propose a learned multi-scale geographic embedding that substantially increases representational capacity.

In summary, we make the following contributions:
\begin{compactitem}
    \item {\bf \ours: Unified Geospatial Contrastive Learning.}  
    We introduce the first purely contrastive framework that jointly aligns an unprecedented set of georeferenced modalities: street-view imagery, aerial imagery, DSMs, text, and geographic coordinates.
    \item {\bf Scaled Latitude--Longitude Encoder.}  
    We propose a novel coordinate encoder that outperforms standard Fourier-feature and MLP-based embeddings by explicitly modeling multi-scale spatial dependencies.
    \item {\bf Strong performance in geospatial tasks.}  
    We demonstrate consistent improvements over single-modality contrastive models and coordinate-only baselines across a diverse suite of geospatial probing and downstream tasks.
\end{compactitem}

\footnotetext{Analytical use of proprietary data sources was done with special permission from Google. Faces and license plates are anonymized by blurring.}

\section{Related Work}

\paragraph{Multimodal Geospatial Models}
Recent developments in geospatial representation learning have transitioned from specialized, task-specific encoders to broader foundation models designed to capture both fine-grained semantic content and large-scale spatial structures \citep{lacoste2023geo}.
Significant focus has been placed on \textit{embedding fields} \citep{brown2025alphaearth, lindenberger2025scaling}, which map geographic coordinates directly to latent features to allow localized spatial interpolation.
Although these fields excel at specific downstream tasks, they essentially act as static snapshots--frozen in time and strictly bound to the geographic distribution of their training sets.
In parallel, multimodal vision models such as \textit{AnySat} \citep{astruc2024anysat}, \textit{Panopticon} \citep{waldmann2025panopticon}, and \textit{Galileo} \citep{tseng2025galileo} have begun to explore the fusion of multiple satellite sensors (\eg, SAR, multispectral).
However, these models combine all sensor observations into a single representation, preventing cross-modal comparison or retrieval.
Moreover, these frameworks remain largely focused on bird's-eye-view data modalities, ignoring the rich ground-level perspective provided by street-level imagery or textual data.

\begin{figure*}
\centering
\begin{tabular}{c@{}c}
\begin{minipage}{0.25\linewidth}
\caption{{\bf Multi-Scale Coordinate Encoder.}
Latitude–longitude coordinates are first projected using multiple random Fourier feature matrices with increasing spectral bandwidths.
Each frequency projection is treated as a token and processed through self-attention to enable inter-scale interaction.
The resulting tokens are averaged to produce a unified $D$-dimensional geographic embedding.}
\end{minipage}
&
\begin{minipage}{0.74\linewidth}
\centering
\resizebox{.95\linewidth}{!}{
\input{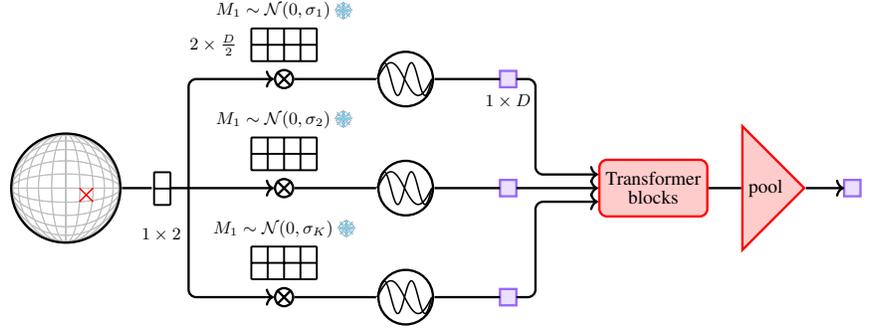}
}
\end{minipage}
\end{tabular}
\end{figure*}

\paragraph{Unified Multimodal Binding and Contrastive Alignment}
The "binding" paradigm seeks to align disparate data streams into a single latent space to support arbitrary modality availability. \textit{ImageBind} \citep{girdhar2023imagebind} popularized the use of a central pivot modality (typically vision) to align sensors, a strategy recently adapted for ecological data in \textit{TaxaBind} \citep{sastry2025taxabind}. In the geospatial domain, contrastive approaches have historically focused on the image-location relationship.
Seminal works like \textit{SatCLIP} \citep{klemmer2025satclip} and \textit{GeoCLIP} \citep{vivanco2023geoclip} established the standard for pairing imagery with geographic coordinates. While these have been refined through improved retrieval functions \citep{dhakal2025range}, temporal dynamics \citep{shatwell2025gt}, and localized attention \citep{liu2025gair}, they typically align only two modalities at a time. Crucially, existing contrastive frameworks largely ignore textual information, despite its central role in modern vision-language models. Unlike \textit{UniBind} \citep{lyu2024unibind}, which still relies on a pivot, our approach adopts a fully holistic all-to-all formulation, ensuring that text, DSM, and visual sensors are all primary citizens in the embedding space.

\paragraph{Geolocation and Cross-Modal Retrieval}
The boundaries of global-scale positioning have been pushed by leveraging the relationship between ground-level and overhead perspectives \citep{Im2gps, Im2gps-pp}. Recent geolocation models like \textit{PIGEON} \citep{haas2023pigeon}, \textit{OpenStreetView} \citep{astruc2024openstreetview} and \textit{Plonk} \citep{dufour2025around} focus on predicting geolocation from street-level image only. Specialized retrieval frameworks, such as \textit{CityLoc} \citep{ma2025cityloc} and \textit{Text2Loc} \citep{xia2024text2loc}, have shown success in urban understanding; however, they are often limited to narrow pairings (e.g., text-to-image or image-to-GPS). ScalingGeoloc \cite{lindenberger2025scaling} aligns Street-View images to aerial image and cell code prototypes. By integrating five modalities simultaneously, our work addresses the limitations of these specialized models, enabling a more robust geospatial reasoning framework that can infer socio-economic variables \citep{sun2024community} or monitor land cover \citep{hu2023mdas} by cross-referencing ground-level, top-down, and elevation data within a single, unified manifold.

\section{Method}

We consider a multimodal sample $x$ characterized by a set of $M$ modalities
$
x = \left\{ x^{1}, x^{2}, \ldots, x^{M} \right\},
$
including street-level imagery (`SV'), aerial imagery (`sat'), elevation models (`DSM'), textual descriptions (`txt'), and geographic coordinates (`GPS').
Our objective is to jointly train modality-specific encoders
$
\left\{ \phi^{m} \right\}_{m \in \mathcal{M}}
$
to output representations that are directly comparable across modalities.

\subsection{Architecture}
Each modality is embedded with a dedicated encoder.

\paragraph{Embedding Earth Observation and Text.}
Street-level and aerial images are encoded with modality-specific image encoders 
$\phi^{\text{SV}}$ and $\phi^{\text{sat}}$, respectively. 
Both are instantiated from the image encoder of SigLIP-2~\citep{tschannen2025siglip}. 
Text inputs are embedded using the SigLIP-2 text encoder $\phi^{\text{txt}}$.
For terrain information, we train a Digital Surface Model (DSM) encoder 
$\phi^{\text{DSM}}$ from scratch. 
This encoder is implemented as a Vision Transformer with register tokens, and use the class token of the last layer as the modality embedding.

\paragraph{Embedding GPS Coordinates.}
We propose a novel learned coordinate encoder $\phi^{\text{GPS}}$ for geographic locations
$x^{\text{GPS}} = \left(x^{\text{lon}}, x^{\text{lat}}\right)$.
To mitigate distortions induced by spherical geometry, we first apply the \emph{Equal Earth Projection}~\cite{vsavrivc2019equal}, mapping latitude--longitude coordinates to a planar representation.
Inspired by GeoCLIP~\cite{tancik2020fourier, vivanco2023geoclip}, we adopt Random Fourier Features to encode spatial information.
We define a random spectral projection matrix $\mathbf{M} \in \mathbb{R}^{\frac{D}{2} \times 2}$ with entries sampled from a Gaussian distribution $\mathcal{N}\left(0, \sigma^2\right)$.
This matrix is fixed and not learned during training.
The encoding
is obtained by concatenating the sine and cosine components:
\begin{align}
    \gamma_{\mathbf{M}}\left(x^{\text{GPS}}\right)
    =
    \left[
    \cos\left(2\pi \mathbf{M} x^{\text{GPS}}\right),
    \sin\left(2\pi \mathbf{M} x^{\text{GPS}}\right)
    \right]^{\top}.
\end{align}

To capture multi-scale spatial structure, we introduce a scalable multi-frequency fusion mechanism.
We perform projections using a set of $K$ matrices $\{\mathbf{M}_k\}_{k=1}^K$, sampled with increasing spectral variances $\{\sigma_k\}_{k=1}^K$.
Each projected embedding is treated as a token.
Unlike prior approaches such as GeoCLIP~\cite{vivanco2023geoclip}, which process each scale independently using separate MLPs and aggregate them by averaging, we explicitly allow interactions across spatial scales.
Specifically, the $K$ tokens are processed by $B$ self-attention blocks with register tokens, enabling information exchange between frequencies.
The final GPS embedding is obtained by averaging the output tokens, yielding a single $D$-dimensional representation.

\begin{figure}
    \centering
    \includegraphics[width=1.0\linewidth]{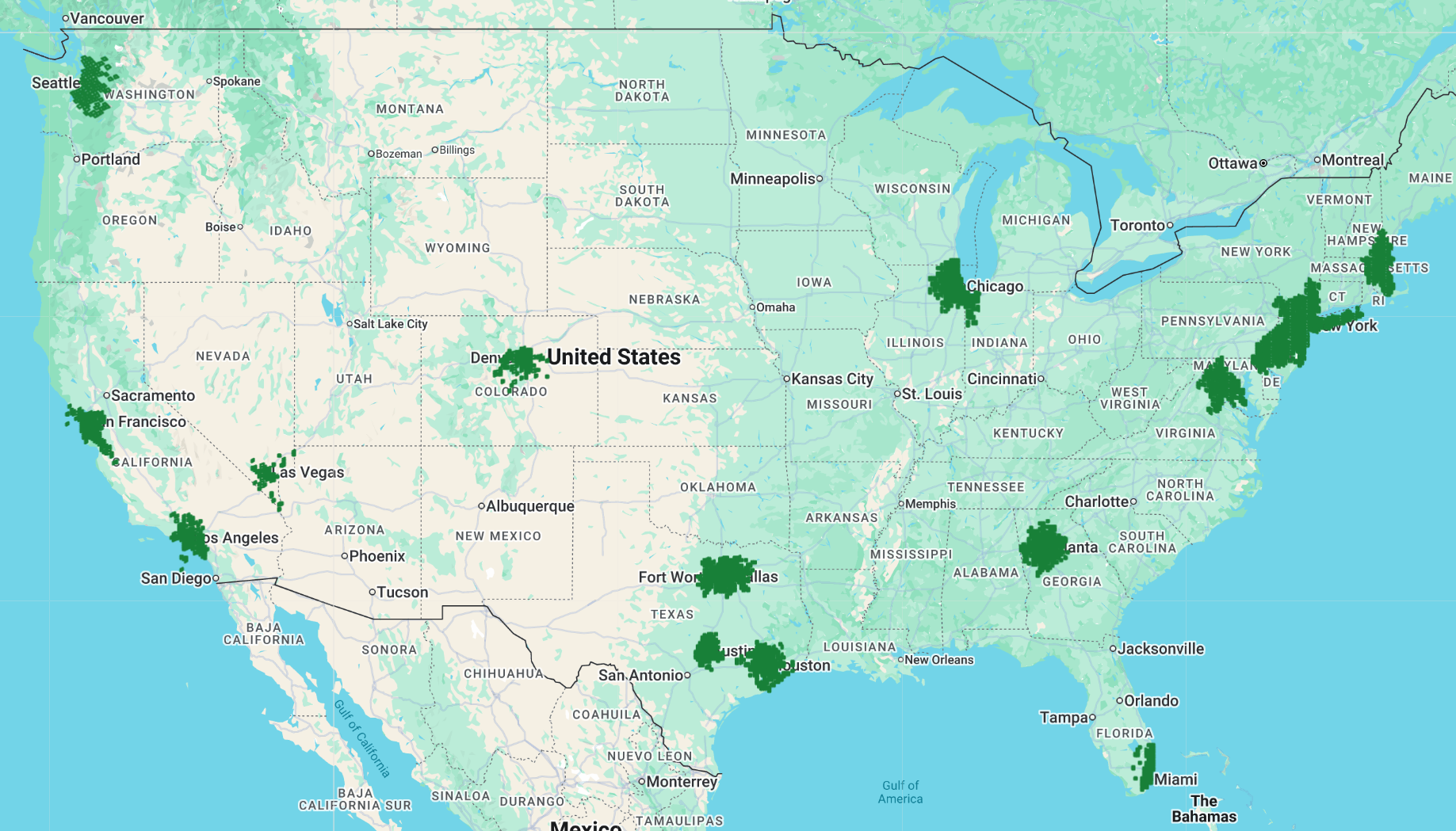}
    \caption{{\bf Geographic Coverage.}
Spatial distribution of sampled locations across the continental United States.
Green regions indicate areas containing samples after spatial filtering and farthest-point sampling.}
    \label{fig:train-coverage}
\end{figure}

\subsection{Supervision}
We consider a batch $\left\{ x_1, \cdots, x_B \right\}$ of multimodal samples, where each sample
$
x_i = \left\{ x_i^{1}, x_i^{2}, \cdots, x_i^{M} \right\}
$
corresponds to a geographic location observed through $M$ co-located modalities.
Each modality is encoded by a modality-specific encoder $\phi^{m}$ into a shared $D$-dimensional embedding
$
f_i^{m} = \phi^{m}\left(x_i^{m}\right).
$
Our objective is to learn embeddings that are \emph{spatially consistent}: representations associated to the same location are close in the embedding space, while those from different locations are far apart.

\paragraph{Multi-Way Contrastive Objective.}
We supervise the encoders using a multi-way contrastive objective that jointly aligns all modalities.
Specifically, we minimize the average InfoNCE loss~\cite{oord2018representation} over all ordered modality pairs $(m,n) \in \mathcal{M}^2$:
\begin{align}
\mathcal{L}
&=
\frac{1}{M^2}
\sum_{(m,n) \in \mathcal{M}^2}
\mathcal{L}_{m \mapsto n},
\\
\mathcal{L}_{m \mapsto n}
&=
-
\frac{1}{B}
\sum_{i=1}^{B}
\log
\left(
    \frac
    {
        \exp
        \left(
        \left\langle
        f_i^{m}, f_i^{n}
        \right\rangle
        /
        \tau
        \right)
    }
    {
        \sum_{j=1}^{B}
        \exp
        \left(
            \left\langle
            f_i^{m}, f_j^{n}
            \right\rangle
            /
            \tau
        \right)
    }
\right),
\end{align}
where $\langle \cdot, \cdot \rangle$ denotes cosine similarity and $\tau$ is a temperature parameter.
By exhaustively contrasting all modality pairs, this objective enforces a fully shared latent space across modalities, enabling robust cross-modal retrieval and reasoning without relying on a privileged pivot modality.

\section{Experiments}
We first describe the dataset used to train \ours~(\cref{sec:dataset}), then present quantitative evaluations on cross-modal retrieval (\cref{sec:retrieval}) and additional downstream tasks (\cref{sec:downtstream}), followed by an ablation study (\cref{sec:ablation}).

\subsection{Dataset}
\label{sec:dataset}

To train and evaluate \ours, we assemble a large-scale multimodal dataset providing a dense, multi-perspective representation of urban environments across the continental United States.

\paragraph{Spatial Extent and Sampling.}
The dataset spans the continental USA, restricted to the largest metropolitan centers, which contain the richest multimodal data.
To ensure uniform spatial coverage, we partition the territory using S2 cells~\cite{s2library}.
We consider cells at level $L=16$, which roughly corresponds to an area of $150\times150$ m.
Our full spatial coverage is composed of $\sim$800k S2 cells.
Within each cell, we apply Farthest Point Sampling~\cite{eldar1997farthest} to select up to 120 street-level panoramas, enforcing a minimum separation of \SI{40}{ \m} between samples.
Cells containing fewer than five valid observations are discarded.
This strategy yields a spatially balanced dataset while avoiding excessive clustering in dense metropolitan areas.

\paragraph{Temporal Split.}
To evaluate robustness under temporal distribution shift, we adopt a strict temporal split.
We run evaluations using data from year 2023, while data from years 2017--2024 (excluding 2023)
is used for training, following the same evaluation setup as Lindenberger \etal~\cite{lindenberger2025scaling}.
This protocol prevents temporal leakage and assesses the ability of the learned representations to generalize across evolving urban landscapes.

\paragraph{Modalities.}
As shown in \cref{fig:dataset}, we collect five complementary data modalities for each geographic location:

\begin{compactitem}
    \item {\bf \sat\ Aerial Imagery.}
    High-resolution overhead imagery is resampled to 60\,cm/pixel resolution and cropped into $256 \times 256$ tiles centered at each location.

    \item {\bf \street\ Street-Level Imagery.}
    Each panorama contributes four perspective crops.
    Panoramic imagery is stitched and rendered using a pinhole camera model, from which $224 \times 224$ crops are sampled with randomized roll, pitch, yaw, and field-of-view to encourage viewpoint robustness.

    \item {\bf \dsm\ Digital Surface Models (DSM).}
    Elevation data provides dense geometric structure aligned with the visual modalities.
    DSM patches are extracted at resolution 60\,cm/pixel and spatially co-registered with the aerial imagery.

    \item {\bf \textual\ Text Descriptions.}
    Each location is associated with an automatically generated textual description derived from large-scale georeferenced data.
    These descriptions capture semantic attributes such as land use, built environment characteristics, and
    context such as local landmarks: see Fig.~\ref{fig:dataset} for an example.

    \item {\bf \gps\ Geographic Coordinates.}
    Raw latitude and longitude corresponding to each sampled location.
\end{compactitem}

\paragraph{Data Sources and Licensing.}
The modalities used in this work are collected from a combination of commercial and proprietary data sources under standard licensing agreements.

\begin{table}[t]
\caption{{\bf Cross-Modal Street View Retrieval.}
We report Acc@100\,m for cross-modal retrieval and specify the modalities contrasted during training. \textsc{OOD} denotes the out-of-domain evaluation setting. \gridcell\ refers to geocells hashcodes.
GeoCLIP is scaled up to the parameter count of our model and retrained on our data.}
    \label{tab:geoloc}
    \centering
    \resizebox{1.0\linewidth}{!}
    {\small

    \setlength{\tabcolsep}{3pt}
    \renewcommand{\arraystretch}{1.0}
    \begin{tabular}{l c cccccc}
    \toprule
         \multirow{2}{*}[-2mm]{\shortstack{retrieval\\[1mm] \street\ \raisedarrow\ \target}}
         & \multirow{2}{*}[-2mm]{\shortstack{training \\ modalities}} & \multicolumn{5}{c}{Target \target } \\\cmidrule(lr){3-7}
          && 
         \sat & \gps & $\{\sat,\gps,\textual,\dsm\}$ & \sat\ \textsc{OOD} &  \gridcell \\
        \midrule
        GeoCLIP \cite{vivanco2023geoclip} & \street\ \gps & - & 24.6 & - & - & \hphantom{4}4.5 \\
        ScalingGeoloc \cite{lindenberger2025scaling} & \street\ \sat\ \gridcell & 45.8 & - & - & - & \bf 56.9 \\ \greyrule
        \bf \ours & \street\ \gps & - & 41.2 & -  & - & 24.8 \\
        \bf \ours & \street\ \sat & 83.9 & - & -  & \bf 41.3 & - \\
        \bf \ours & \street\ \sat\ \gps & 76.7 & 46.5 &  75.6 & 32.3 &  29.0 \\
        \bf \ours & \street\ \sat\ \gps\ \dsm& 77.2 & 47.0 & 81.9 & 33.5 & 29.6 \\
        \bf \ours & \street\ \sat\ \gps\ \dsm\ \textual & \bf 88.2 & \bf 69.4 & \bf 91.0 &  \bf 41.2 & 41.2 \\

       \bottomrule
    \end{tabular}
}
\end{table}

\subsection{Cross-Modality Retrieval}
\label{sec:retrieval}

We evaluate cross-modal alignment through a zero-shot geospatial retrieval task, with results reported in \cref{tab:geoloc}.
Given a street-view query, the objective is to retrieve the geographically matching instance in another modality.
Performance on this task quantifies the consistency of semantic and spatial representations across modalities.

\paragraph{Evaluation Protocol.}
Following the standard retrieval-based localization paradigm, we $\ell_2$-normalize all embeddings and compute cosine similarities between a {\em query modality}
and a {\em database of georeferenced candidates}. 
We choose ground-level images \street\ as queries, and use modalities in $\{\sat,\gps,\textual,\dsm\}$ as targets.
The predicted location corresponds to the candidate with the highest similarity score. To ensure a fair comparison with GeoCLIP, we fine-tuned it on our training set, scaled its coordinate encoder to match our parameter count, and utilized an identical training regime. 
We evaluate street-view queries against the following targets:

\begin{compactitem}
    \item {\bf Aerial \sat.}
    This corresponds to the classic cross-view retrieval task, matching ground-level imagery to overhead observations.

    \item {\bf GPS Coordinates \gps.}
    This setting evaluates direct image-to-location retrieval~\cite{Im2gps}, where street-view embeddings are matched against encoded geographic coordinates.

    \item {\bf Multimodal Ensembling $\{\sat,\gps,\textual,\dsm\}$.}
    To assess complementarity across modalities, we compute separate similarity matrices for each available modality and aggregate them by simple averaging before selecting the global maximum.
    This measures the synergy between heterogeneous geospatial signals. In Table~\ref{tab:geoloc}, this indicates ensembling the data modalities each model was trained with (excluding ground-level images).

    \item {\bf Out-of-Distribution Aerial \sat\ (OOD).}
    We evaluate cross-view retrieval on a geographically distinct region (Amsterdam) to assess spatial generalization beyond the training distribution (USA).

    \item {\bf Geocells \gridcell.}
    Following the protocol of Scaling Geoloc~\cite{lindenberger2025scaling}, we evaluate localization via spatial discretization.
    The study area is partitioned into discrete cells, each represented by encoding the centroid coordinates using our latitude--longitude encoder.
    Since our model is not explicitly trained on geocell tokens, this setting probes its ability to generalize to discretized spatial representations.
\end{compactitem}

\begin{table}[t]
\caption{{\bf Satellite Image Encoder.}
We evaluate the ability of our satellite image encoder to analyze aerial and satellite imagery of various image encoders on two geospatial benchmarks: solar panel detection \panel\ and land-cover segmentation \landcover.
Models are grouped into cross-modal contrastive frameworks (top), Earth observation foundation models (middle), and general-purpose vision foundation models (bottom).
We \underline{underline} the best performance among contrastive models and \textbf{bold} the best overall performance.}
    \label{tab:aerial}
    \centering
    \resizebox{1.0\linewidth}{!}
    {\small
    \setlength{\tabcolsep}{3pt}
    \renewcommand{\arraystretch}{1.0}
    \begin{tabular}{l c c ccc}
    \toprule
        \multirow{2}{*}{\makecell{classif / semseg \\ \sat\ \raisedarrow\  \target}} &\multirow{2}{*}{\makecell{training\\modalities}} & & m-pv4ger \panel & m-chesapeake \landcover \\
          & & model & classif (OA) & semseg (mIoU) \\
        \midrule    
        SatClip \cite{klemmer2025satclip}   &\sat\ \gps\ & ViT-B & 93.0	& 59.3\\
        SigLip2 \citep{tschannen2025siglip} &\sat\ \textual &ViT-B & 95.7 & 60.9\\
        \bf \ours & \sat\ \gps\   & ViT-B & 96.9 & 65.9 \\
        \bf \ours &\street\ \sat\ \gps\ \dsm\ \textual   & ViT-B & \underline{97.0}	& \underline{66.3}\\\greyrule 
         SenPaMAE \citep{prexl2024senpa} & & ViT-B & 87.1 & 46.9\\
        DOFA \citep{xiong2024neural} && ViT-L & 97.4 & 59.2\\
        AnySat \citep{astruc2024anysat} && ViT-B & 92.8 & 61.7\\
        Panopticon \citep{waldmann2025panopticon} && ViT-B$^\star$ & 96.4 & 60.8\\\greyrule 
        Dinov2 \citep{oquab2023dinov2} && ViT-B$^\star$  & 97.5 & 64.0\\
        Dinov3 Web \citep{simeoni2025dinov3}  & & ViT-7B & \bf 98.3 & \bf 76.5\\
        
       \bottomrule
    \end{tabular}
}
\end{table}

\paragraph{Analysis.}
From the results in \cref{tab:geoloc}, we draw the following conclusions:
\begin{compactitem}
    \item \textbf{Stronger Cross-Modal Alignment.}
    \textsc{UniGeoCLIP} consistently outperforms a retrained GeoCLIP model~\cite{vivanco2023geoclip} for retrieval-based geolocation. This indicates improved cross-modal harmonization.

    \item \textbf{Impact of Multi-Contrastive Formulation.}
    Our models improve the more modalities we contrast. Note that contrasting with text requires a larger batch size to converge.

    \item \textbf{Complementarity of Modalities.}
    Multimodal ensembling consistently surpasses the best individual modality. Ground-level imagery captures fine-grained semantics, whereas aerial imagery and DSMs encode structural and spatial layout cues; their combination yields a more robust and discriminative retrieval capacity.

    \item \textbf{Generalization to Spatial Discretization.}
    Under the geocell protocol, GeoCLIP struggles to generalize. In contrast, our model achieves competitive localization accuracy despite never being trained for cell-based classification, indicating greater spatial flexibility.

    \item \textbf{Out of Distribution.}
    We show that our model can generalize to unseen areas with different statistics, by evaluating on a city in Europe (Amsterdam) using a model trained in the USA.
\end{compactitem}

\begin{table}[t]
  \caption{{\bf Evaluation of the Location Encoder.}
Performance on 27 downstream regression tasks spanning health, socio-economic, and environmental indicators.
Models are grouped into contrastive approaches (top) and pre-computed embedding fields (bottom).
Variants marked with $\dag$ are retrained on our dataset.
$\star$ denotes models specifically trained for socio-economic prediction.
We \textbf{bold} the best overall performance and \underline{underline} the best performance among contrastive approaches.}
    \label{tab:latlng}
    \centering
    \resizebox{1.0\linewidth}{!}
    {\small
    \begin{tabular}{l ccc c}
    \toprule
         \multirow{2}{*}[-2mm]{\makecell{Regression \\ \gps\ \raisedarrow\ \target}} & \multicolumn{3}{c}{target \target} \\ \cmidrule(lr){2-4}
         &health & social & environmental & mean R² \\ 
        \midrule
        SigLIP 2 \citep{tschannen2025siglip} & 34.5 & 48.5 & 72.3 & 40.2\\
        SatClip \citep{klemmer2025satclip} & 25.7 & 34.4 & 70.7 & 30.1\\
        $\dag$ SatClip \citep{klemmer2025satclip} & 32.3 & 48.5 & 43.7 & 36.7\\
         GeoClip \citep{vivanco2023geoclip} & 45.2 & 65.7 & \underline{\bf 84.2} & 49.8\\
        $\dag$ GeoClip \citep{vivanco2023geoclip} & 47.3 & 67.9 & 77.9 & 51.6\\
        \bf \ours & \underline{53.1} & \underline{69.9} & 81.1 & \underline{57.0}\\
       \greyrule
        AlphaEarth \citep{brown2025alphaearth} & 23.1 & 29.9 & 83.3 & 29.0\\
        $\star$ PDFM \citep{agarwal2024general} & \bf 73.9 & \bf 82.6 & 82.3 & \bf 74.5 \\
       \bottomrule
    \end{tabular}
}
\end{table}

\begin{figure*}[t]
    \centering
    \sbox{\tempbox}{\includegraphics[width=0.3\textwidth, trim={40cm 2cm 40cm 2cm}, clip]{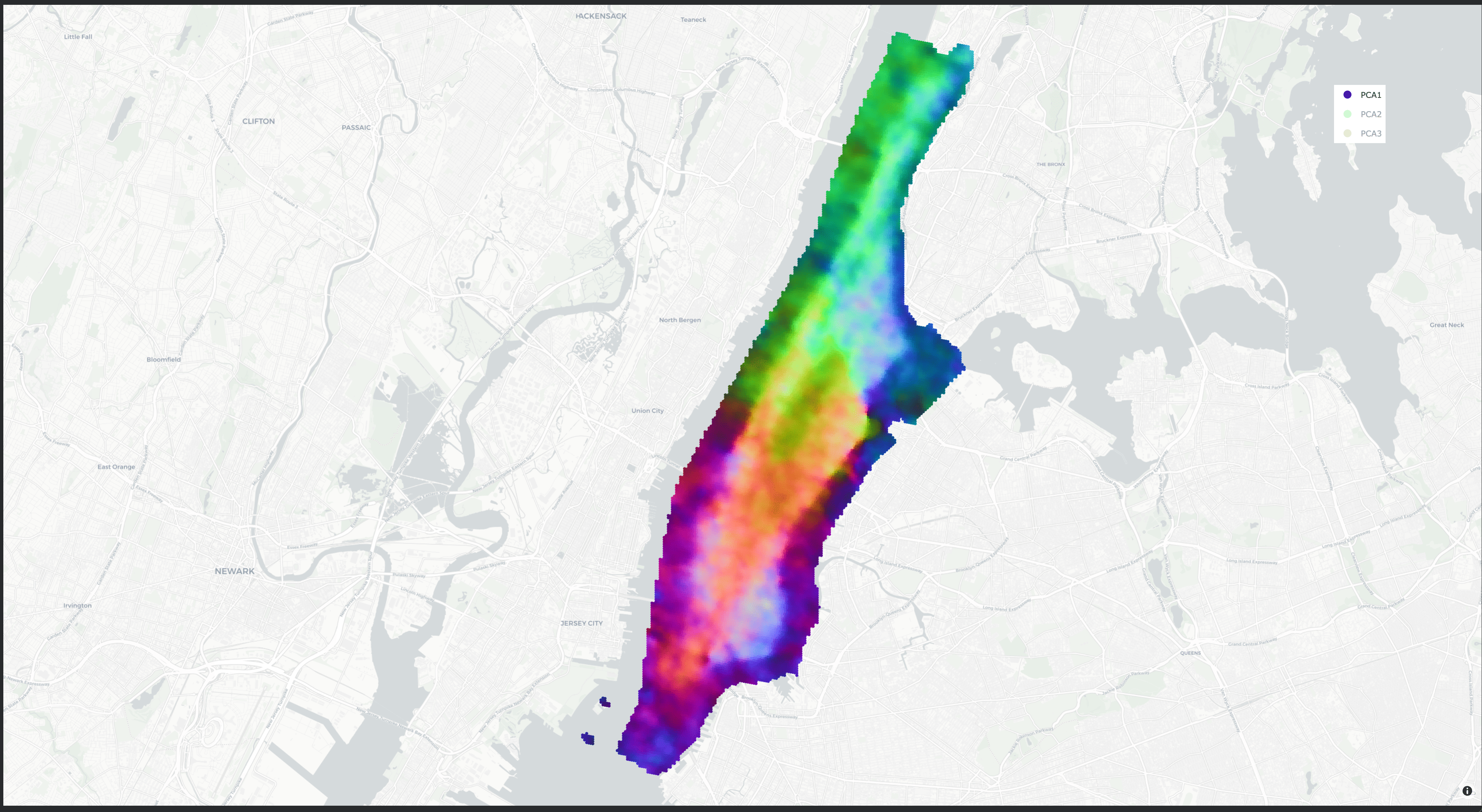}}
    \begin{subfigure}[b]{0.3\textwidth}
        \centering
        \includegraphics[width=\textwidth, trim={40cm 2cm 40cm 2cm}, clip]{images/pca_latlng/manhattan_full_1.png}
        \caption{UniGeoClip (ours)}
        \label{fig:image1}
    \end{subfigure}
    \hfill
    \begin{subfigure}[b]{0.3\textwidth}
        \centering
        \includegraphics[height=\ht\tempbox]{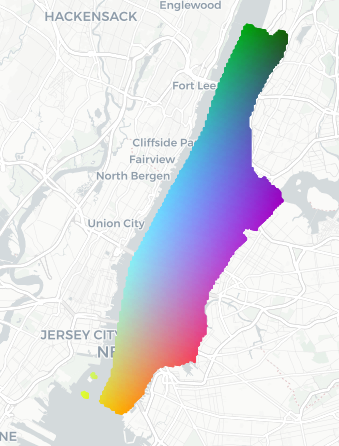}
        \caption{SatCLIP}
        \label{fig:image2}
    \end{subfigure}
    \hfill
    \begin{subfigure}[b]{0.3\textwidth}
        \centering
        \includegraphics[height=\ht\tempbox]{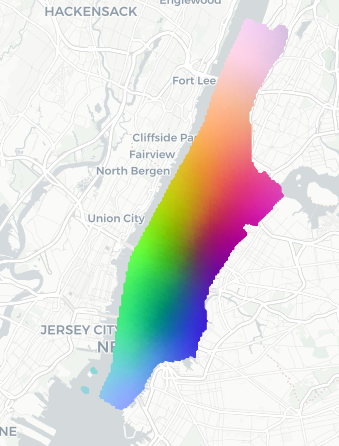}
        \caption{GeoCLIP}
        \label{fig:image3}
    \end{subfigure}
    
\caption{{\bf PCA of Coordinate Embeddings.}
Embeddings computed over a dense grid in Manhattan, NYC are projected using PCA, with the top three principal components mapped to RGB.
UniGeoCLIP produces spatial patterns that reflect underlying urban structure (e.g., Central Park and surrounding neighborhoods), indicating semantically informed representations.
In contrast, SatCLIP and GeoCLIP exhibit smoother, predominantly position-driven gradients. }
    \label{fig:PCA-ll}
\end{figure*}

\subsection{Encoder Downstream Evaluation}
\label{sec:downtstream}
We evaluate the capacity of individual encoders trained with our framework to generalize to downstream tasks.
Encoders are frozen and assessed via linear probing.

\paragraph{Satellite Image Encoder.}
We evaluate the aerial encoder on two complementary geospatial tasks: 
(i) \textit{m-pv4ger} \cite{lacoste2023geo,mayer20223pv4ger} is a photovoltaic panel detection benchmark based on high-resolution aerial imagery, requiring the identification of small structures within large scenes;
(ii) \textit{m-chesapeake} \cite{lacoste2023geo,robinson2019chesa} is a land-cover semantic segmentation dataset requiring dense pixel-wise classification.
For segmentation, we extract patch embeddings from the final transformer layer and project them to class logits of size $P{\times}P{\times}C$, where $P$ denotes the patch size and $C$ the number of classes.
Together, these tasks assess both global semantic discrimination and fine-grained spatial reasoning.

As shown in \cref{tab:aerial}, \textbf{UniGeoCLIP} achieves the strongest performance among all contrastive frameworks.
It surpasses both \textit{SigLIP2} and a fine-tuned version of \textit{SatCLIP}~\cite{klemmer2025satclip} on m-pv4ger (97.0\% OA) and m-chesapeake (66.3 mIoU), highlighting the representational advantages of multimodal geospatial pre-training.

On semantic segmentation, our model also outperforms specialized Earth observation models such as \textit{AnySat} and \textit{Panopticon}, despite never been trained for dense tasks.
Although large-scale general-purpose models such as \textit{DINOv3 Web}~\cite{simeoni2025dinov3} achieve higher absolute performance, they rely on substantially larger architectures (e.g., ViT-7B) and web-scale pre-training.

Overall, these results suggest that the multi-modal contrastive objective effectively distills structural geographic information into the aerial encoder, enabling strong downstream performance without task-specific supervision or increased model scale.

\paragraph{Spatial Coordinate Encoder.}
We evaluate the representation power of our novel coordinate encoder on the 27 downstream regression tasks introduced by Sun \etal~\cite{sun2024community}, where geographic coordinates are mapped to health, socio-economic, and environmental indicators.
We limit this evaluation to locations that overlap with our training set.
This yields 1,447 training locations and 179 test locations.

We then perform linear probing on our frozen coordinate embeddings and compare against several baselines:
(i)~other contrastive models, both off-the-shelf and retrained on our dataset,
(ii)~the text encoder \textit{SigLIP 2}~\cite{tschannen2025siglip}, with which the geographic information is provided via a detailed textual prompt that includes the country, state, county, city, and geographic coordinates,
(iii)~pre-computed embedding fields inferred by \textit{AlphaEarth}~\cite{brown2025alphaearth} from Earth observations by \textit{PDFM}~\cite{agarwal2024general} from rich auxiliary socio-economic and environmental covariates.

\Cref{tab:latlng} shows that our encoder achieves a mean $R^2$ of 57.0, substantially outperforming existing embedding fields from \textit{AlphaEarth} (29.0) and contrastive baselines like \textit{SatCLIP} (30.1) and \textit{GeoCLIP} (49.8).
Importantly, even when compared to our retrained versions of these baselines under identical data conditions, our coordinate encoder exhibits stronger spatial generalization.

While \textit{PDFM} remains state-of-the-art on this regression-focused benchmark, it leverages a substantially broader set of auxiliary spatial signals—including thematic maps, search trends, and environmental covariates.
In contrast, our approach relies solely on multimodal alignment of fundamental geospatial inputs.
These results indicate that contrastive multi-modal training induces rich spatial representations that correlate strongly with socio-economic and environmental indicators.

Retraining SatCLIP and GeoCLIP on our dataset significantly improves performance in health and social regression tasks. This gain is likely due to the spatial alignment between our training locations and the benchmark, allowing the models to capture localized socio-economic nuances. Conversely, environmental performance remains stagnant or decreases; these tasks likely benefit more from the vast geographic diversity of the original models' broader training sets, which capture large-scale ecological patterns that our metropolitan-focused data may lack.

We visualize in \cref{fig:PCA-ll} the representations produced by our coordinate encoder by applying Principal Component Analysis (PCA) to embeddings computed over a dense grid of locations in Manhattan, New York.
The resulting projection reveals a continuous and semantically structured spatial representation.
Rather than discretely hashing geographic regions, the encoder learns a smooth manifold characterized by coherent clusters and gradual transitions that reflect underlying urban structure, such as Central Park.

Compared to alternative spatial encodings, our embeddings exhibit sharper boundaries and more distinctive spatial organization, indicating improved representational fidelity and multi-scale geographic modeling.

\paragraph{Evaluation of the DSM Encoder.}
We consider the task of pixelwise land-cover semantic segmentation of DSM images from the \textbf{MDAS} dataset~\cite{hu2023mdas}, 
which contains 1,702 high-resolution images totaling 24G annotated pixels.
Unlike aerial imagery, DSM data currently lacks widely adopted large-scale foundation models.
We therefore compare our DSM encoder against two standard baselines trained from scratch: a U-Net and a Vision Transformer (ViT).
In contrast, our DSM encoder is evaluated under a linear probing protocol with frozen weights.

As shown in \cref{tab:dsm}, our model outperforms both baselines by a substantial margin.
By aligning elevation data with semantic text and visual imagery during pre-training, the encoder acquires structurally informed representations that standard architectures trained solely with semantic supervision fail to capture.
These results indicate that cross-modal contrastive pre-training serves as an effective initialization strategy for DSM understanding, compensating for the absence of domain-specific pre-trained models and yielding a more robust backbone for elevation-driven tasks.

\begin{table}[t]
    \centering
    \begin{tabular}{cc}
\begin{minipage}{0.48\linewidth}
   \caption{{\bf Evaluation of the DSM Encoder on MDAS.}
We evaluate the accuracy of land-cover prediction from DSM-only input.
Our linearly-probed DSM encoder significantly outperforms U-Net and ViT-B models trained from scratch.}
    \label{tab:dsm}
    \end{minipage}
    &
    \begin{minipage}{0.52\linewidth}
        \vspace{-10mm}
     {\footnotesize
    \begin{tabular}{l c ccc}
    \toprule
         \dsm\ \raisedarrow\ \landcover & accuracy \\
        \midrule
        U-Net \citep{ronneberger2015unet} & 45.5\\
        ViT-B \citep{dosovitskiy2020vit} & 39.1\\
       \greyrule
       \bf \ours & \bf 72.0 \\
       \bottomrule
    \end{tabular}
}
    \end{minipage}
    \end{tabular}
\end{table}
\subsection{Discussion and Ablation Study}
\label{sec:ablation}

We present an ablation study to analyze the impact of key design choices and an analysis to better understand the behavior of our multimodal alignment framework.

\paragraph{Which Modalities Contribute Most?}
To quantify the contribution of each modality to the shared embedding space, we evaluate retrieval performance under different training configurations.
As shown in \cref{tab:geoloc}, performance improves consistently as additional modalities are incorporated.
A model trained only on aerial imagery and coordinates already achieves competitive zero-shot localization.
However, adding Street View significantly enhances spatial precision, and the full multimodal configuration (including text and DSM) reaches a peak Accuracy@\SI{100}{ \m} of 69.4\%.
These results confirm that each modality contributes complementary spatial information.

\begin{figure}
    \centering
    \input{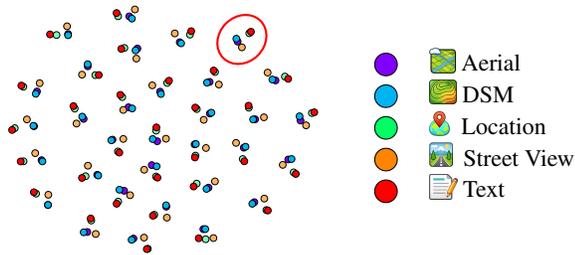}
    \caption{{\bf Location Embedding Visualization.}
    T-SNE projection of embeddings for 48 distinct locations.
    Each cluster \protect\tikz[scale=0.4]{\protect\draw[thick, red] (0,0) circle(2mm);} corresponds to a single geographic location and contains embeddings from all modalities.
    }\label{fig:tsne}%
\end{figure}

\paragraph{Can We Align Heterogeneous Modalities?}
Our framework contrasts modalities that are fundamentally different in nature: text, imagery, elevation data, and raw geographic coordinates.
Although these modalities vary in structure and dimensionality, they share a common latent factor: the geographic content of a location.
We thus wonder whether such heterogeneous signals are really embedded into a shared space without sacrificing representational expressivity.
To investigate this, we visualize in \cref{fig:tsne} the learned latent space using t-SNE \cite{van2008visualizing}.

Embeddings of distinct modalities form compact clusters corresponding to individual geographic locations.
The consistent co-localization of all five modalities within these clusters indicates that the model learns a modality-invariant geographic representation, rather than segregating modalities into disjoint subspaces.
Within each cluster, embeddings of aerial and DSM are generally closer to each other, as are those of location and street view.

\paragraph{Should We Scale the Location Encoder?}
The location encoder operates on low-dimensional inputs consisting of latitude and longitude.
This raises the question of whether increasing architectural capacity is justified for mapping two scalar values into a rich embedding space.
To examine the relationship between model capacity and representational power, we vary the number of self-attention blocks in the GPS encoder and evaluate performance across multiple retrieval tasks.
As reported in \cref{tab:ablations}, increasing the depth of the encoder yields consistent gains across all cross-modal retrieval settings, including aerial and out-of-distribution (OOD) evaluation.

At \textit{Depth 0}, the encoder reduces to fixed Random Fourier Features without transformer layers.
In this regime, performance is limited: coordinate retrieval reaches only 10.2, and cell-code generalization drops to 4.4.
As depth increases, retrieval accuracy improves steadily.
With 12 self-attention blocks, the coordinate retrieval rises to 47.0, and multimodal ensembling reaches 79.1.
These results suggest that, despite the low dimensionality of its input, the coordinate encoder benefits substantially from increased capacity.
The transformer layers enable non-linear interactions across spatial frequencies, allowing the model to capture complex geographic structure beyond simple spectral projections.

\begin{table}[t]
    \caption{{\bf Ablation of the number of blocks of the location encoder.}
    This ablation is conducted with lower batch size.}
    \label{tab:ablations}
    \centering
    {\footnotesize
\begin{tabular}{l c cccc}
\toprule
     Retrieval
     &  \multicolumn{5}{c}{Target \target } \\\cmidrule(lr){2-6}
      \street\ \raisedarrow\ \target & 
     \sat & \gps & $\{\sat,\gps,\textual,\dsm\}$ & \sat\ \textsc{OOD} & \gridcell \\
    \midrule
    Depth 0 & 58.1 & 55.0 & 10.2  & 13.6 & 4.4 \\
    Depth 4 & 77.2 & 73.1 & 40.7  & 30.0 & 25.4 \\
    Depth 8 & 77.6 & 73.1 & 44.0  & 27.8 & 27.6 \\
   \greyrule
   Depth 12 & 79.1 & 74.4 & 47.0  & 29.2 & 29.3\\
   \bottomrule
\end{tabular}
}
\end{table}

\paragraph{Comparison with Alternative Location Encoding Paradigms.}
To evaluate the impact of our multi-scale design, we replace our coordinate encoder with two existing alternatives scaled to match our parameter count: \textit{GeoCLIP}~\cite{vivanco2023geoclip}, using Random Fourier Features~\cite{tancik2020fourier}, and \textit{SatCLIP}~\cite{klemmer2025satclip}, using SIREN~\cite{russwurm2023geographic}.
All models are trained using identical modalities and supervision, for a fair comparison.

Despite parity in capacity, our approach significantly outperforms the scaled baselines on coordinate regression tasks (\cref{tab:latlng}) and also improves downstream aerial representation quality (\cref{tab:aerial}).
In particular, our joint aerial–coordinate pre-training consistently exceeds the performance of the scaled \textit{SatCLIP} baseline.
Our multi-scale frequency fusion mechanism leverages depth to model complex geographic dependencies through cross-scale interactions that shallow MLP- or RFF-based encoders fail to capture.

\begin{table}[t]
    \caption{{\bf Comparison of different location encoders.}
    Our model is trained on same data modalities as the baselines.}
    \label{tab:ablations_encoder}
    \centering
    {\footnotesize
\setlength{\tabcolsep}{3pt}
\begin{tabular}{l cccc}
    \toprule
    \makecell[l]{location encoder\\architecture} & \makecell{location\\retrieval} & \makecell{geocell\\ retrieval} & \makecell{aerial \\ semseg} & \makecell{socio-economic \\ variables}\\ 
    \midrule
    SirenNet (SatCLIP) & - & - & 59.3 & 35.5 \\
    RFF encoder (GeoCLIP) & 24.6 & 4.5 & - & 52.6 \\
   \greyrule
   \bf Our location encoder & {\bf 41.2} & {\bf 24.8} & {\bf 65.9} & {\bf 56.7} \\
   \bottomrule
\end{tabular}
}
\end{table}

\paragraph{Generalization Beyond the Training Region.}
To assess the generalization capacity of our learned representations, we perform zero-shot geolocation in an out-of-distribution (OOD) setting.
We curate an evaluation dataset from Amsterdam, a region excluded from pre-training and exhibiting a substantial domain shift relative to the training data (U.S. cities).
The dataset contains aligned aerial imagery, GPS coordinates, and street-level observations.

Localization is performed via cross-modal retrieval: Street View queries are matched against an aerial image gallery, and the latitude–longitude of the top-retrieved aerial image is used as the predicted location.
As reported in \cref{tab:geoloc}, the model trained solely with aerial and Street View contrast achieves the strongest OOD performance.
This is expected, as incorporating additional modalities may encourage specialization to the geographic distribution of the training data.
Nevertheless, the full multimodal model maintains competitive performance, demonstrating that the learned embeddings retain strong generalization beyond the training region.
Importantly, performance trends in this OOD evaluation remain consistent with those observed in the primary in-distribution experiments.
This stability across geographic domains highlights the robustness of our multimodal alignment strategy.

\section{Conclusion}

We introduced \ours, a unified multimodal framework for global geographic representation learning.
By adopting a multimodal, all-to-all contrastive objective, our method is able to align five heterogeneous modalities (aerial imagery, Street View, coordinates, DSM, and text) within a shared embedding space.
This holistic alignment improves zero-shot geolocation, cross-modal retrieval, and downstream encoder performance.
We also propose a transformer-based coordinate encoder that outperforms traditional spatial encoder baselines.
Overall, \ours\ provides a scalable foundation for multimodal geospatial reasoning and large-scale Earth observation applications.

\ARXIV{
The authors thank Simon Lynen and Filip Saina for inspiring discussions and valuable feedback on the manuscript.
We thank Mimi Sun for helping with the evaluation of PDFM and Matthew Pereira for helping with the text summaries.
}{}

\FloatBarrier
\pagebreak
\pagebreak
\section*{Acknowledgments}

The authors thank Simon Lynen and Filip Saina for inspiring discussions and valuable feedback on the manuscript.
We thank Mimi Sun for helping with the evaluation of PDFM and Matthew Pereira for helping with the text summaries.

{\small
\balance{\bibliographystyle{templates/EV/ieeenat_fullname}
\bibliography{mybib}}
}

\ARXIV{
    \FloatBarrier
    \pagebreak
    \balance
    \section*{\centering \LARGE Appendix}
    \setcounter{section}{0}
    \setcounter{figure}{0}
    \setcounter{table}{0}
    \renewcommand*{\theHsection}{appendix.\the\value{section}}
    \renewcommand\thefigure{\arabic{figure}}
    \renewcommand\thetable{\arabic{table}}
    \renewcommand\thefigure{A-\arabic{figure}}
\renewcommand\thesection{A-\arabic{section}}
\renewcommand\thetable{A-\arabic{table}}
\renewcommand\theequation{A-\arabic{equation}}
\renewcommand\thealgorithm{A-\arabic{algorithm}}

}{}

\end{document}